\documentclass[conference]{IEEEtran}
\usepackage[utf8]{inputenc}
\usepackage{cite}
\usepackage{algorithm}
\usepackage{algorithmic}
\usepackage{graphics}
\usepackage{graphicx}
\graphicspath{ {./images/} }

\title{PlantTracing: Tracing Arabidopsis Thaliana Apex with CenterTrack}
\author{
\IEEEauthorblockN{Yuanzhe Liu}
\IEEEauthorblockA{
\textit{New York University}\\
yl7897@nyu.edu}
\and
\IEEEauthorblockN{Yixiang Mao}
\IEEEauthorblockA{
\textit{New York University}\\
yixiang.mao@nyu.edu}
\and
\IEEEauthorblockN{Yao Wang}
\IEEEauthorblockA{
\textit{New York University}\\
yw523@nyu.edu}
}
\date{December 2021}

\begin{document}

\maketitle

\begin{abstract}

This work applies an encoder-decoder-based machine learning network to detect and track the motion and growth of the flowering stem apex of Arabidopsis Thaliana. Based on the CenterTrack, a machine learning back-end network, we trained a model based on ten time-lapsed labeled videos and tested against three videos. 
\end{abstract}

\section{Introduction}

The motion of plants has been a popular topic in botany. Research shows that plants can adjust their morphology to absorb the energy from nature. This phenomenon makes tracking the apex movement meaningful in studying how plants adapt to the environment. Today, we can record the plant's pose via a high-resolution camera, generating a sequence of time-lapsed pictures or videos. Previously, the apex position is measured manually. However, we can automate this process via machine learning. We can use videos with manually marked center positions and bounding boxes as input and train the network using GPUs. We can then use this network to automatically detect the apex, measure the center of the apex, and track its movement. Previous work \cite{Plant-Tracer} adopts block matching algorithm \cite{block-matching} to track Arabidopsis seedlings apex from side-view in a time-lapse video. However, there are several limitations exist. Users have to select the apex manually and set the center of the apex location since this method cannot detect the apex. In addition, the method cannot track and associate the apex object in continuous frames. Finally, if the tracking fails, the method cannot correct itself. We propose a CNN-based solution to first detect the apex object and associate objects in different frames to track them to resolve the above issue. This project implements CenterTrack~\cite{CenterTrack} that can detect and track the pre-trained objects developed based on CenterNet\cite{CenterNet}, which uses the center position of bounding box to represent an object using auto encoder-decoder model~\cite{encoderdecoder} and Deformable Convolution Neural Network\cite{deformableCNN}. The CenterTrack takes the current and previous frames as training inputs to predict the object's center position in the next frame. The detected result of the next frame is then associated with the prediction from the previous two frames to associate the last object and the current object to achieve tracking. The model is retrained with a dataset of plants in which each video frame is labeled with a bounding box for 90 epochs on Nvidia GPU 1080 for three days. 

\section{Background}
\subsection{Basic}

Auto encoder-decoder~\cite{encoderdecoder} is a Recurrent Neural Network (RNN). The input is first encoded into states represented by a sequence of vectors and then decoded to generate the output vectors. RNN improves from CNN by introducing feedback paths that utilize the result of the current or next layer as input along with the input from the previous layer. 

\begin{figure}[h]
    \includegraphics[width=\linewidth]{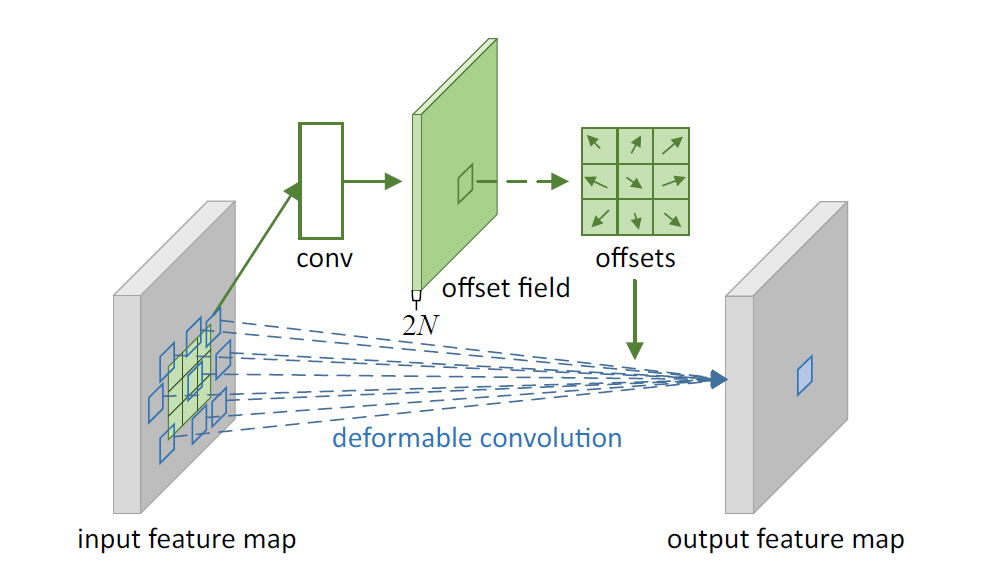}
    \caption{Architect of Deep Layer Aggregation from \cite{deformableCNN}}
    \label{fig:pattern}
\end{figure}

The Deformable Convolution Networks\cite{deformableCNN}, or deformable ConvNets, adds a 2D offset to the fixed grid locations in convolution. The output feature map now gets the result from the previous input feature map through a fully connected layer containing a 2D offset field after the convolution. 

\begin{figure}[h]
    \includegraphics[width=\linewidth]{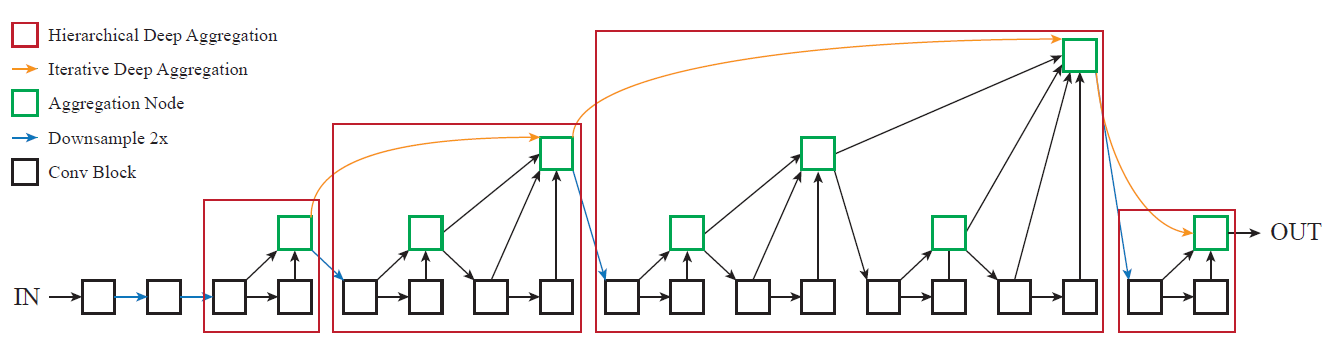}
    \caption{Architect of Deep Layer Aggregation\cite{DLA}}
    \label{fig:pattern}
\end{figure}

Deep Layer Aggregation\cite{DLA} is a fully convolutional encoder-decoder network that can fuse information across layers by attractively and hierarchically merging the features in different layers to improve accuracy with fewer parameters.

\subsection{Model}

\begin{figure}[h]
    \includegraphics[width=\linewidth]{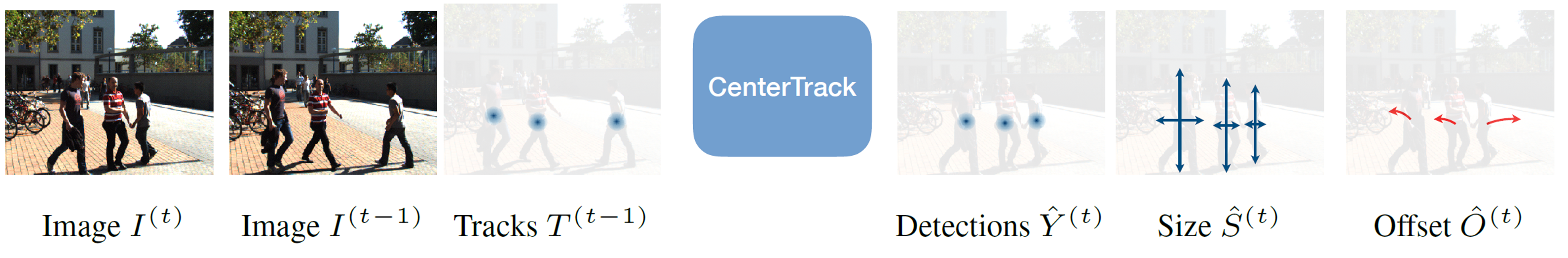}
    \caption{CenterTrack input and output\cite{CenterTrack}}
    \label{fig:pattern}
\end{figure}

CenterNet takes a single 3-channel image $I \in R^{W \times H \times 3}$ and produces $(p_i, s_i)$ where p is the center point coordinate $(x, y)$ and s is the height and width of the object's bounding box from a low-resolution heat map $\hat{Y} \in [0, 1]^{\frac{W}{R} \times \frac{H}{R} \times C}$ where R = 4 is the down-sampling factor and a size map $\hat{S} \in R^{\frac{W}{R} \times \frac{H}{R} \times 2}$. The CenterTrack takes the image from previous frame $I^{(t - 1)} \in R^{W \times H \times 3}$ and current frame $I^{(t)} \in R^{W \times H \times 3}$ as well as the tracked objects in previous frame $T^{t-1} = \{b_0^{t-1}, b_0^{t-2}, ...\}_i$ where $b = (p, s, w, id)$. W is the detection confidence where $w \in [0, 1]$. and id is the unique id allocated to each object detected. In addition to the output from CenterNet, a channel of 2D displacement $\hat{D}^t \in R^{\frac{W}{R} \times \frac{H}{R} \times 2}$ is generated that contains the difference in location of the object between the current frame and previous frame. The Greedy algorithm associates the detected object's center position with the predicted center position calculated from the previous position and the offset for each object. A new tracklet is generated if there is no matching.

\section{Experiments}

\begin{figure}[h]
    \includegraphics[width=\linewidth]{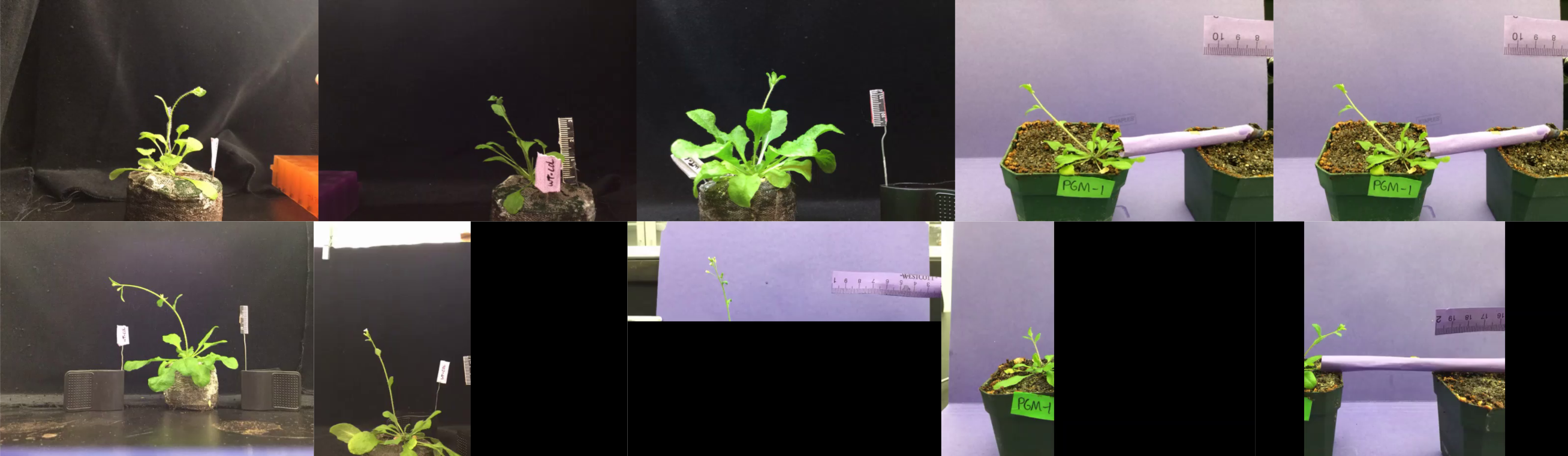}
    \caption{Training video setup}
    \label{fig:pattern}
\end{figure}

\subsection{Dataset}

Time-lapse videos are taken with smartphones mounted on tripods every two minutes, and images are encoded into a video with 30 fps. Each video can last for 10 hours to up to three days. An example setup is shown in Figure 1 for ten training videos. The setup has a solid white or black background, a ruler placed at the same focal length of the plant as a reference to indicate the scale for measurement, and some labels of the experiment. The videos are first split into frames and manually labeled using the LabelIMG. The output is then converted from Pascal VOC XML format to COCO JSON format as the input annotation files to the network. Video sequences 1, 2, 3, 4, 5, 7, 8, 10, 11, and 12 are the training input while sequences 6, 21, 9 are used for testing. 

\begin{table}[h]
    \centering
    \begin{tabular}{|c|c|c|c|c|c|c|c|c|c|c|}
        \hline
         Video & 1 & 2 & 3 & 4 & 5 & 7 & 8 & 10 & 11 & 12 \\
        \hline
        Frames & 1426 & 1777 & 919 & 1064 & 1047 & 490 & 1126 & 977 & 1064 & 1064 \\ 
        \hline
    \end{tabular}
    \vspace{-1em}
    \caption{Number of Frames in each training video.} 
\end{table}

\begin{table}[h]
    \centering
    \begin{tabular}{|c|c|c|c|c|}
        \hline
         Video & 6 & 9 & 21 & 22 \\
        \hline
        Frames & 1042 & 1126 & 2426 & 1431 \\ 
        \hline
    \end{tabular}
    \vspace{-1em}
    \caption{Number of Frames in each testing video.} 
\end{table}

\subsection{Implementation}
\begin{figure}[h]
    \includegraphics[width=\linewidth]{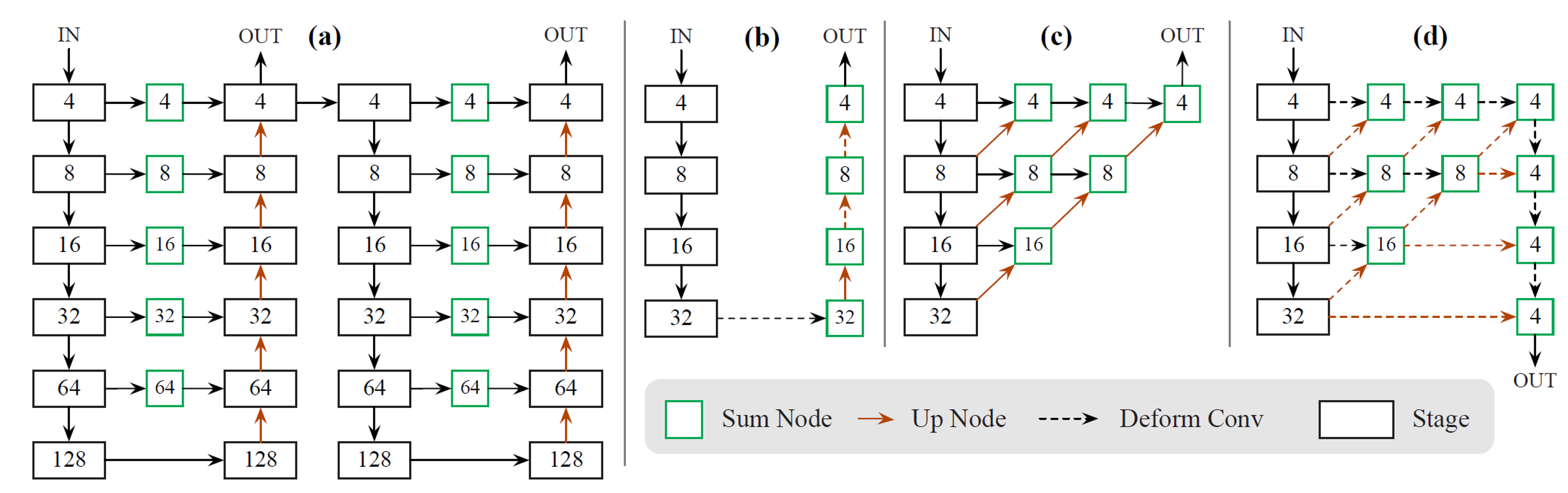}
    \caption{Options of Backbone used in CenterNet\cite{CenterNet}}
    \label{fig:pattern}
\end{figure}
DLA-34\cite{DLA} is used as the backbone of the network in figure 3c using Adam optimization\cite{kingma2017adam} with a learning rate 1.25e-4 and batch size 5. Since detection is usually used in the lab, no augmentations are performed on the training dataset. The network is trained for 90 epochs on a machine with an Intel i7-8700K CPU and a Nvidia 1080 GPU for three days. 

\subsection{Result}
\begin{table}[h]
    \centering
    \begin{tabular}{|c|c|c|c|c|c|}
        \hline
        Video & Center MSE & Failed & More Object & Total Frames\\
        \hline
        6 & 3.391382431 & 0 & 0 & 1042 \\
        \hline
        9 & 2.12492227 & 79 & 0 & 1126 \\
        \hline
        21 & 6.992375844 & 397 & 339 & 2426 \\
        \hline
        22 & 7.212677611 & 175 & 85 & 1431 \\
        \hline
    \end{tabular}
    \vspace{-1em}
    \caption{Number of Frames in each training video.} 
\end{table}

\begin{table}[b]
    \centering
    \begin{tabular}{|c|c|c|c|c|c|}
        \hline
        Center MSE & Failed & More Object & Total Frames & Tracking Threshold \\
        \hline
        2.25301509 & 0 & 755 & 1126 & 0.3 \\
        \hline
        2.12492227 & 79 & 0 & 1126 & 0.4 \\
        \hline
    \end{tabular}
    \vspace{-1em}
    \caption{Result of Video 9 for different tracking Threshold} 
\end{table}

\begin{figure}[h]
    \includegraphics[width=\linewidth]{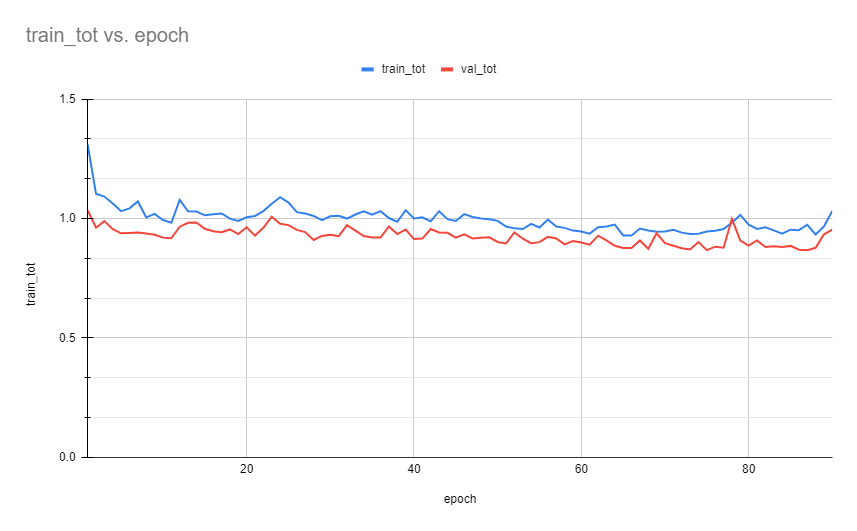}
    \caption{Training and Validation Loss}
    \label{fig:pattern}
\end{figure}

We chose video sequences 6, 9, 21, and 22 to match the result in \cite{Plant-Tracer}, and the mean square error (MSE) is used to measure the difference between the actual position and the prediction result utilizing the weight of epoch 87 which has a training loss of 0.974175 and validation loss of 0.867303. Since this work approaches the tracking by first detecting the apex in the video and then associating objects in contiguous frames, the number of frames the network failed to detect, and the number of frames where more than one object is detected is recorded. Since CenterTrack won't render the object if its coincidence is below a certain threshold, we can also tweak that threshold to improve the MSE.

\begin{figure}[h]
    \includegraphics[width=\linewidth]{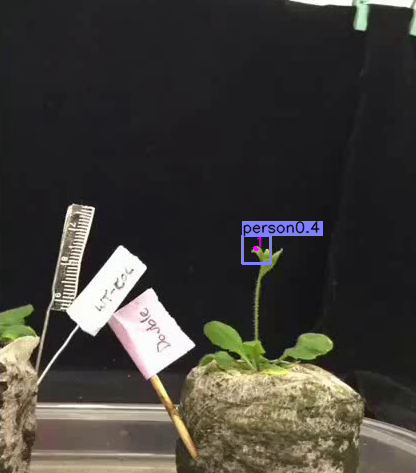}
    \caption{Video 21 detection result}
    \label{fig:pattern}
\end{figure}

\begin{figure}[h]
    \includegraphics[width=\linewidth]{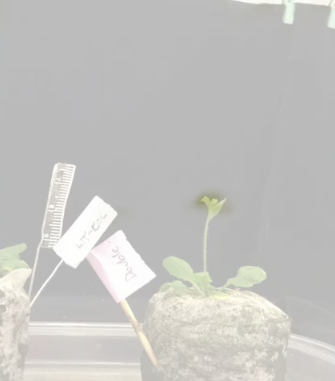}
    \caption{Video 21 tracking heat map}
    \label{fig:pattern}
\end{figure}

\begin{figure}[h]
    \includegraphics[width=\linewidth]{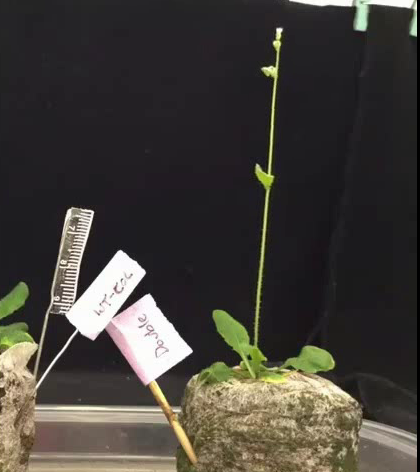}
    \caption{Video 21 no apex detected}
    \label{fig:pattern}
\end{figure}

\begin{figure}[h]
    \includegraphics[width=\linewidth]{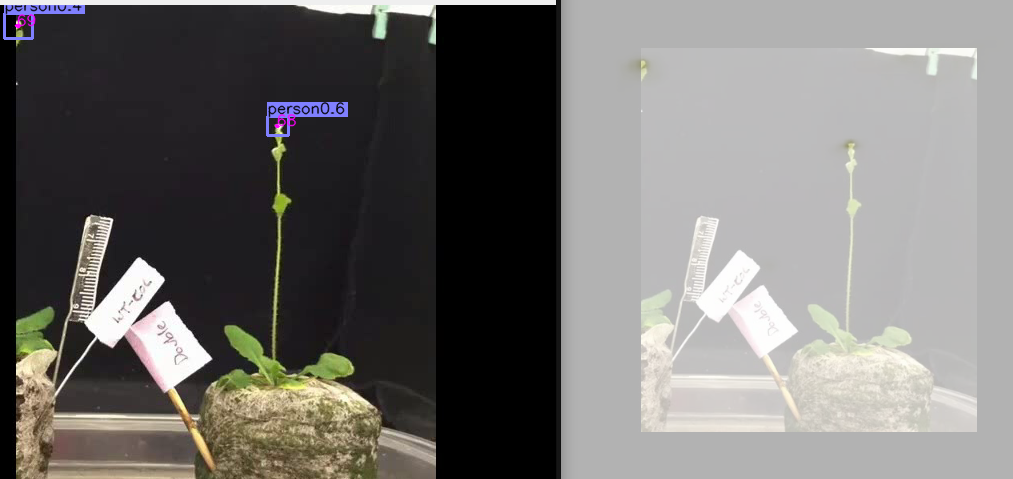}
    \caption{Video 22 with 2 apex detected trace}
    \label{fig:pattern}
\end{figure}

\subsection{Analyze}

The result shows that the overall mean square error is better than previous work using U-Net with post-processing and the KLT method. However, since the network needs to detect the object first if no apex is detected or its confidence is below the threshold, the network can't associate and, therefore, track the apex object, as illustrated in video 21 in Figure 9. In video 6, the network can detect the object in every frame, and the error is small. However, that's not the case for testing videos 21 and 22 since the apex shows relatively large movement and shape changes. In addition, there are 339 frames in video 21, and 85 frames in video 22 have more than one object detected. In most cases, that's an invalid result since we only have 1 plant in the video, so only one apex object should be detected. However, in Video 22, we can see a second apex of another plant was accidentally being captured, showing this network can successfully detect it

We can improve the result by increasing the tracking threshold. The default is set to 0.3. In Video 9, we observe that the corner of a piece of paper used for labeling the experiment is accidentally identified as an apex. Since CenterTrack is designed to track multiple objects and associate them according to the detection result, increasing the tracking threshold of the object confidence will eliminate false positives since the network won't treat the object as a detected result. By increasing the tracking threshold for video 9 from 0.3 to 0.4, we observe that no frames contain more than one object, but we have 79 more frames, and the network failed to detect any apex. However, this is acceptable since there are 1126 frames. 

Future improvement will focus on adding more datasets and reworking part of the architecture as well as adding an LSTM module to utilize info from all previous frames to reduce the number of frames that the network failed to detect or the number of frames in which the network contains more than one apex objects in which one of them is not an apex at all. 

\section{Conclusion}

In this project, I implement CenterTrack, an auto-encoder-decoder network enhanced with deep layer aggregation and deformable CNN to track the apex of Arabidopsis Thaliana. Results show improvement over previous methods using KLT and U-Net. 

\bibliographystyle{IEEEtran}
\bibliography{ref}

\appendix
\section{Detection Image}
\begin{figure}[h]
    \includegraphics[width=\linewidth]{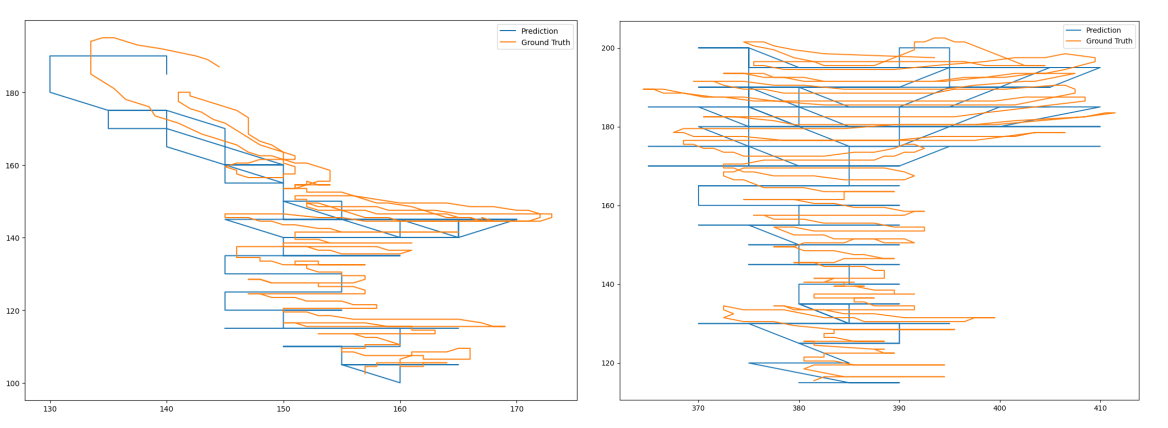}
    \caption{Video 6 and 9 trace}
    \label{fig:pattern}
\end{figure}

\begin{figure}[h]
    \includegraphics[width=\linewidth]{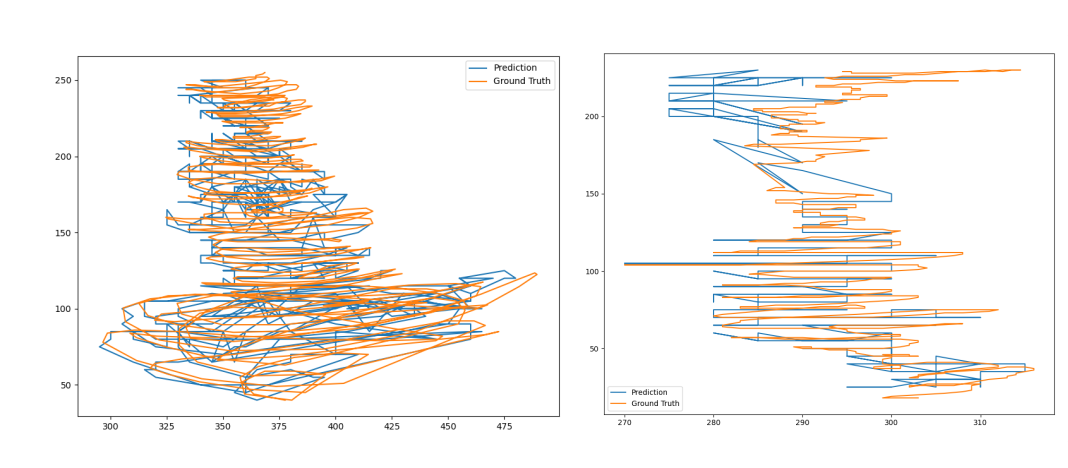}
    \caption{Video 21 and 22 trace}
    \label{fig:pattern}
\end{figure}

\begin{figure}[h]
    \includegraphics[width=\linewidth]{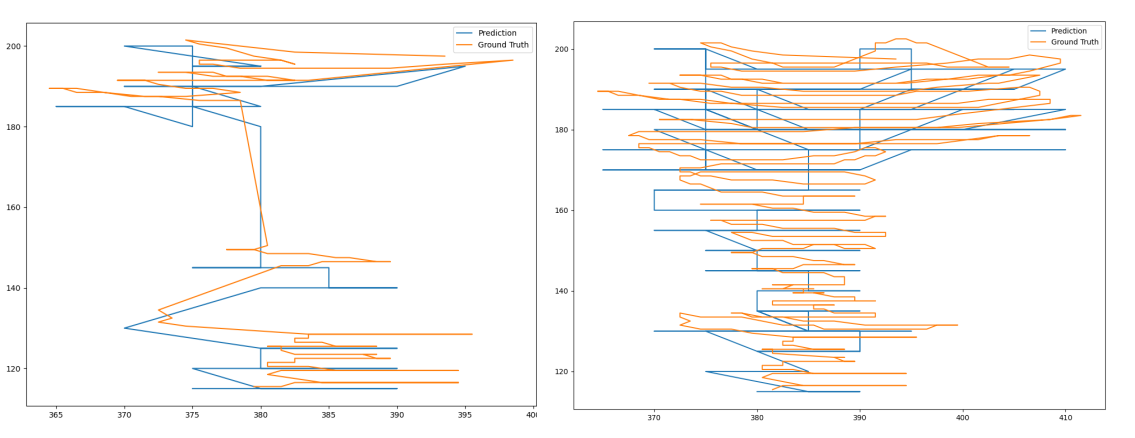}
    \caption{Video 9 trace}
    \label{fig:pattern}
\end{figure}
\end{document}